\documentclass[10pt,twocolumn,letterpaper]{article}

\usepackage[pagenumbers]{paperstyle}

\usepackage{graphicx}
\usepackage{amsmath}
\usepackage{amssymb}
\usepackage{booktabs}
\usepackage{multirow}
\usepackage{xcolor}
\usepackage{tikz}
\usepackage{algorithm}
\usepackage{algpseudocode}
\usetikzlibrary{arrows.meta,positioning,shapes.geometric}

\definecolor{linkblue}{rgb}{0.21,0.49,0.74}
\usepackage[pagebackref,breaklinks,colorlinks,allcolors=linkblue]{hyperref}
\usepackage[capitalize]{cleveref}
\crefname{section}{Sec.}{Secs.}
\Crefname{section}{Section}{Sections}
\crefname{table}{Tab.}{Tabs.}
\crefname{algorithm}{Alg.}{Algs.}
\Crefname{algorithm}{Algorithm}{Algorithms}
\Crefname{table}{Table}{Tables}

\title{Smart Surveillance: Deep Learning and Hybrid Approaches for\\Dynamic Scene Analysis, Object Detection and Motion Tracking}

\author{Shahran Rahman Alve,  Shahnewaz Siddique \\
Department of Electrical and Computer Engineering, North South University\\
Bashundhara R/A, Dhaka 1229, Bangladesh\\
{\tt\small shahran.alve@northsouth.edu,shahnewaz.siddique@northsouth.edu}}

\begin{document}
\maketitle

\begin{abstract}
Event-triggered recording reduces the footage a surveillance camera must store, but the trigger decides how much idle video is kept, how many events are missed, and how much computation is spent. We compare six motion triggers (frame differencing, dense optical flow, MOG2, KNN, CNT, GSOC), a CNN detector, strided detection, and three motion-gated hybrids under one recording policy, on a benchmark of seven static-camera scenes (5,800 frames, 27 annotated events) with thresholds transferred to unseen scenes by leave-one-video-out. Detector-based triggers stored 19--28\% of idle frames, against 36--89\% for motion triggers. Under simulated camera jitter, frame differencing, optical flow, MOG2 and KNN stored 97--100\% of idle frames, while detector triggers stayed below 29\%. Running the detector on every 8th frame kept 99.7\% of event frames at 9.8\,ms of CPU per frame, cheaper and more robust than gating it with background subtraction. On VisDrone, raising YOLO11n's input from 640 to 1280\,px raised AP from 15.5 to 26.8, more than a larger model, and INT8 quantization cut latency by 17\% for 0.3 AP. In tracking, longer track memory raised IDF1 from 70.5 to 76.7, but every tracker recovered only 39--44\% of overlapped people, because detection rather than association failed. Extended experiments show that an adaptive-stride trigger, which checks an empty scene every 16th frame and an occupied one every 8th, cut detector calls to 10.2\% of frames while storing fewer idle frames (26.4\%); that a 1\,s instead of 2\,s post-roll cut total storage by 3 points without losing event frames; that YOLO26s at 1280\,px reached 34.8 AP while test-time augmentation lowered AP; and that with three frames of coasting a Kalman tracker lost only 1.4 MOTA when 30\% of detections were dropped. Code, annotations and raw results are available from the author.
\end{abstract}

\section{Introduction}
\label{sec:intro}

Most surveillance cameras record continuously, so much of what they store shows an empty scene. Recording only when something happens saves storage and review time, but a frame that is not recorded can never be reviewed. The core design question is which signal should trigger recording, and what that choice costs in missed events, wasted storage and computation.

Two families of triggers are common. Motion cues such as frame differencing and background subtraction~\cite{stauffer1999adaptive,zivkovic2004improved,zivkovic2006efficient} are cheap but react to anything that changes pixels, including lighting and camera shake. CNN detectors~\cite{ren2015faster,liu2016ssd,redmon2016yolo,ge2021yolox,yolo11} react to people and vehicles but cost far more per frame, and their confidence scores may not transfer across scenes. Video-analytics systems combine the two, using cheap filters before an expensive model~\cite{kang2017noscope,li2020reducto}, yet these options are rarely compared on the same footage with the same recording policy.

We measure them, asking four questions. \textbf{RQ1:} How much idle footage does each trigger store when tuned to keep at least 95\% of event frames in every scene? \textbf{RQ2:} How do triggers degrade under illumination change, camera jitter and sensor noise? \textbf{RQ3:} Which detector and input size give the best accuracy per CPU millisecond on elevated views with small objects, and how much does INT8 quantization help? \textbf{RQ4:} How much do a motion model and longer track memory help multi-object tracking when people overlap? \textbf{RQ5:} Can the detector trigger be made cheaper without storing more idle video, by adapting its stride, fusing cues, filtering motion or changing the recording buffer? \textbf{RQ6:} How well does tracking survive when detections arrive only on some frames or are missed at random?

Our contributions are:
(1) an event-recording benchmark of 5,800 frames from seven static-camera scenes with 27 annotated events, a shared recording policy, and a leave-one-video-out protocol that tests whether a threshold transfers to an unseen scene;
(2) a controlled comparison of six motion triggers, a detector, strided detection and three gated hybrids on clean video and under three seeded perturbations;
(3) a CPU-only detector study on VisDrone covering three YOLO generations, YOLOX, input resolution, tiled inference and INT8 quantization;
(4) a tracking study that isolates the Kalman motion model and track memory and reports recall separately for overlapped people;
(5) an adaptive-stride trigger and an analysis of strides, cue fusion, persistence filtering, buffer length, threshold transfer, recall-target sensitivity, reaction time and per-scene behaviour;
(6) a study of tracking with strided and randomly missing detections, with and without coasting, and of the tracker's sensitivity to its parameters.
The recording policy, threshold selection, adaptive trigger and sparse-detection tracker are given as \cref{alg:record,alg:lovo,alg:adaptive,alg:sparse}.

\section{Related Work}
\label{sec:related}

\paragraph{Background subtraction and motion.}
Adaptive per-pixel Gaussian mixtures~\cite{stauffer1999adaptive} remain the standard way to separate moving objects from a static scene; MOG2 chooses the number of components per pixel~\cite{zivkovic2004improved}, and KNN estimates the background density non-parametrically~\cite{zivkovic2006efficient}. Sample-based models such as ViBe~\cite{barnich2011vibe} trade the statistical model for speed. CDnet 2014~\cite{wang2014cdnet} showed that these methods still fail under illumination change, jitter and dynamic backgrounds. Dense~\cite{farneback2003two} and sparse~\cite{lucas1981iterative} optical flow give a second motion cue, and qualitative motion segmentation~\cite{bouthemy1993motion} extends it to moving cameras. Surveys catalogue dozens of such models~\cite{bouwmans2014traditional}, and the CDnet benchmarks~\cite{goyette2012changedetection,wang2014cdnet} drove methods with adaptive feedback such as SuBSENSE~\cite{stcharles2015subsense}. Deep background subtraction now leads these benchmarks: FgSegNet~\cite{lim2018fgsegnet} learns multiscale foreground features per scene, and BSUV-Net~\cite{tezcan2020bsuv} targets unseen videos, the setting our leave-one-video-out protocol imitates, but both need a GPU at video rate. Hybrids of optical flow and deep features have been used for activity recognition~\cite{tanberk2020hybrid} and unsafe-behaviour detection~\cite{ding2018deep}, and event cameras sense brightness changes directly in hardware~\cite{gallego2022event}.

\paragraph{Object detection.}
Two-stage~\cite{ren2015faster} and single-stage~\cite{liu2016ssd,redmon2016yolo} detectors define the speed-accuracy trade-off; later single-stage models include YOLOX~\cite{ge2021yolox} and the YOLOv8, YOLO11 and YOLO26 releases~\cite{yolov8,yolo11,yolo26}, while transformer detectors~\cite{carion2020detr} are costly on CPUs. Multi-scale features~\cite{lin2017fpn}, the focal loss~\cite{lin2017focal} and compound scaling~\cite{tan2020efficientdet} improved accuracy per FLOP, and lightweight backbones~\cite{howard2017mobilenets}, distillation~\cite{hinton2015distilling} and integer quantization~\cite{jacob2018quantization} target edge devices. Most detectors are trained on COCO~\cite{lin2014coco}, whose objects are larger than those seen by elevated cameras; VisDrone~\cite{zhu2021visdrone} provides elevated views with many small people and vehicles, and methods for it add transformer heads~\cite{zhu2021tph} or slice the image into tiles~\cite{akyon2022sahi}, which we compare against simply enlarging the input.

\paragraph{Tracking.}
SORT~\cite{bewley2016sort} combines a Kalman filter~\cite{kalman1960new} with Hungarian assignment; DeepSORT~\cite{wojke2017deepsort} adds appearance; the IOU tracker~\cite{bochinski2017iou} uses overlap alone; ByteTrack~\cite{zhang2022bytetrack}, OC-SORT~\cite{cao2023ocsort} and BoT-SORT~\cite{aharon2022botsort} improve association under occlusion. Joint detection-and-tracking models regress boxes from the previous frame (Tracktor~\cite{bergmann2019tracktor}), predict object centres and offsets (CenterTrack~\cite{zhou2020centertrack}) or share one network for detection and re-identification (FairMOT~\cite{zhang2021fairmot}), and StrongSORT~\cite{du2023strongsort} revisits DeepSORT with stronger components. Standard metrics are MOTA~\cite{bernardin2008clear}, IDF1~\cite{ristani2016idf1} and HOTA~\cite{luiten2021hota}, usually on MOTChallenge~\cite{lealtaixe2015motchallenge,milan2016mot16,dendorfer2020mot20}. Single-object trackers include Siamese~\cite{bertinetto2016siamfc,li2018siamrpn}, correlation-filter~\cite{bolme2010mosse,henriques2015kcf,lukezic2017csrdcf} and online-learning~\cite{babenko2011mil} methods, evaluated with the OTB protocol~\cite{wu2013otb}.

\paragraph{Frame filtering for video analytics.}
NoScope~\cite{kang2017noscope} skips frames with difference detectors and specialised models before a full CNN; Chameleon~\cite{jiang2018chameleon} adapts frame rate and model configuration; Focus~\cite{hsieh2018focus} indexes video cheaply at ingest; Reducto~\cite{li2020reducto} filters frames on the camera with low-level features. VideoStorm~\cite{zhang2017videostorm} trades accuracy for delay across many queries, and FilterForward~\cite{canel2019filterforward} runs cheap per-application filters on constrained edge nodes, the setting described as the ``killer app'' for edge computing~\cite{ananthanarayanan2017real}. Anomaly detection addresses a different question, namely which recorded events are unusual, with weak supervision~\cite{sultani2018real}, sparse reconstruction~\cite{lu2013abnormal} or future-frame prediction~\cite{liu2018future}. These systems report savings on query workloads; we instead ask which trigger keeps idle footage out of storage without losing events, and how robust that choice is to lighting and shake.

\section{Background}
\label{sec:background}

\paragraph{Background subtraction.}
A Gaussian-mixture background model~\cite{stauffer1999adaptive} describes the history of each pixel $x$ by $K$ weighted Gaussians, $p(\mathbf{I}_t(x))=\sum_{k=1}^{K}w_k\,\mathcal{N}(\mathbf{I}_t(x);\boldsymbol{\mu}_k,\sigma_k^2\mathbf{I})$. A pixel is foreground if no component with enough weight explains it within a few standard deviations; matched components are updated with learning rate $\alpha$, $\boldsymbol{\mu}_k\leftarrow(1-\alpha)\boldsymbol{\mu}_k+\alpha\mathbf{I}_t(x)$, so that a stationary object is absorbed into the background after roughly $1/\alpha$ frames. MOG2~\cite{zivkovic2004improved} chooses $K$ per pixel, KNN~\cite{zivkovic2006efficient} replaces the mixture by a kernel density over recent samples, and CNT and GSOC are faster counting-based and sample-based variants~\cite{opencv}. All of them respond to any pixel change, which is why global gain changes and camera shake (\cref{sec:results}) defeat them.

\paragraph{Optical flow.}
Dense flow assumes brightness constancy, $\mathbf{I}_t(x)=\mathbf{I}_{t+1}(x+\mathbf{u})$; Farneb\"ack's method~\cite{farneback2003two} fits a quadratic polynomial to each neighbourhood and solves for the displacement $\mathbf{u}$ that aligns the two expansions. A flow trigger thresholds the fraction of pixels with $\|\mathbf{u}\|$ above a magnitude, so it measures motion rather than difference.

\paragraph{Single-stage detection.}
YOLO-family detectors~\cite{redmon2016yolo,ge2021yolox,yolo11} predict, for every cell of several feature maps, class scores and box offsets in one forward pass and remove duplicates with non-maximum suppression. Their cost grows roughly with the number of input pixels, which is why input size, not only model size, sets the speed--accuracy trade-off for small objects (\cref{sec:results}).

\paragraph{Kalman tracking.}
SORT~\cite{bewley2016sort} represents a track by $\mathbf{x}=[u,v,s,r,\dot u,\dot v,\dot s]^\top$ (box centre, area, aspect ratio and their velocities) and uses a constant-velocity Kalman filter~\cite{kalman1960new}: the prediction $\hat{\mathbf{x}}_{t|t-1}=\mathbf{F}\hat{\mathbf{x}}_{t-1}$, $\mathbf{P}_{t|t-1}=\mathbf{F}\mathbf{P}_{t-1}\mathbf{F}^\top+\mathbf{Q}$ is corrected by a matched detection $\mathbf{z}_t$ through the gain $\mathbf{K}_t=\mathbf{P}_{t|t-1}\mathbf{H}^\top(\mathbf{H}\mathbf{P}_{t|t-1}\mathbf{H}^\top+\mathbf{R})^{-1}$. Detections are assigned to predictions by the Hungarian algorithm on IoU. Without detections, the prediction step alone extrapolates the box, which is what lets a tracker bridge strided or missed detections (RQ6).

\paragraph{Metrics.}
For tracking, $\mathrm{MOTA}=1-\sum_t(\mathrm{FN}_t+\mathrm{FP}_t+\mathrm{IDSW}_t)/\sum_t\mathrm{GT}_t$~\cite{bernardin2008clear} counts misses, false positives and identity switches, while $\mathrm{IDF1}=2\,\mathrm{IDTP}/(2\,\mathrm{IDTP}+\mathrm{IDFP}+\mathrm{IDFN})$~\cite{ristani2016idf1} measures how long each person keeps one identity. For detection we use COCO-style AP~\cite{lin2014coco}, averaged over IoU thresholds 0.5:0.95 and broken down by object size. Single-object trackers are ranked by the area under the OTB success curve~\cite{wu2013otb}.

\section{Triggers and Recording Policy}
\label{sec:method}

\begin{figure}[t]
\centering
\includegraphics[width=\columnwidth]{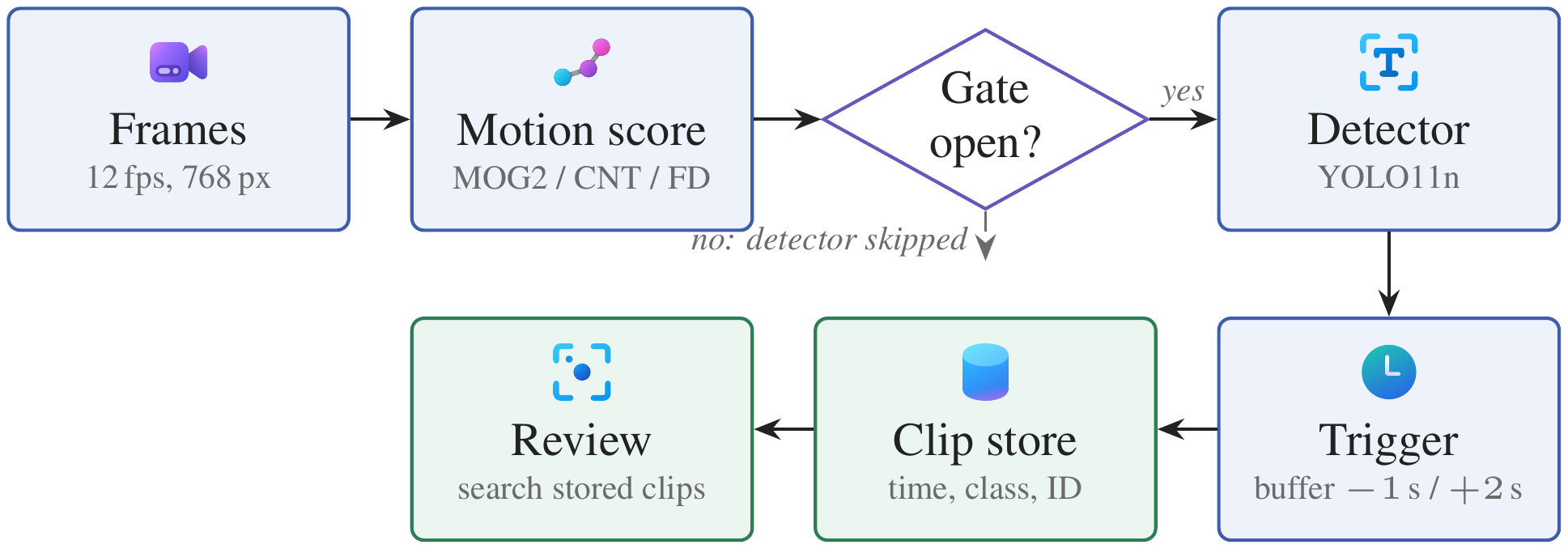}
\caption{Gated-detector pipeline. Single-cue triggers drop the gate (detector only) or the detector (motion only); the strided trigger replaces the gate with ``every $k$-th frame''.}
\label{fig:pipeline}
\end{figure}

Every trigger produces a per-frame score; a threshold turns it into a binary trigger, and one shared policy turns triggers into stored clips (\cref{fig:pipeline}).

\begin{algorithm}[t]
\caption{Event-triggered recording with a ring buffer}
\label{alg:record}
\begin{algorithmic}[1]
\Require frames $I_1,I_2,\dots$; trigger $\textsc{Trig}$; rate $r$; pre-roll $a$\,s; post-roll $b$\,s
\State $B\gets$ ring buffer of capacity $\lceil a r\rceil$ frames; $t_{\text{last}}\gets-\infty$
\For{$t=1,2,\dots$}
  \If{$\textsc{Trig}(I_t)$}
    \If{$t-t_{\text{last}}>\lceil b r\rceil$}
      \State \textsc{Write}$(B)$; empty $B$ \Comment{pre-roll}
    \EndIf
    \State $t_{\text{last}}\gets t$
  \EndIf
  \If{$t-t_{\text{last}}\le\lceil b r\rceil$}
    \State \textsc{Write}$(I_t)$ \Comment{event or post-roll}
  \Else
    \State $B.\textsc{Push}(I_t)$
  \EndIf
\EndFor
\end{algorithmic}
\end{algorithm}

\begin{algorithm}[t]
\caption{Leave-one-video-out threshold selection}
\label{alg:lovo}
\begin{algorithmic}[1]
\Require videos $\mathcal{V}$; candidate thresholds $\Theta$; recall target $\rho=0.95$
\For{each held-out video $v\in\mathcal{V}$}
  \State $\mathcal{T}\gets\mathcal{V}\setminus\{v\}$
  \State $\Theta^{+}\gets\{\theta\in\Theta:\ \textsc{Recall}(\theta,u)\ge\rho\ \ \forall u\in\mathcal{T}\}$
  \If{$\Theta^{+}\neq\emptyset$} $\theta_v\gets\arg\min_{\theta\in\Theta^{+}}\textsc{Stored}(\theta,\mathcal{T})$
  \Else\ $\theta_v\gets\arg\max_{\theta\in\Theta}\min_{u\in\mathcal{T}}\textsc{Recall}(\theta,u)$
  \EndIf
  \State evaluate $\theta_v$ on $v$ (clean and perturbed copies)
\EndFor
\end{algorithmic}
\end{algorithm}

\begin{algorithm}[t]
\caption{Adaptive-stride detector trigger}
\label{alg:adaptive}
\begin{algorithmic}[1]
\Require detector score $s(\cdot)$; threshold $\theta$; idle stride $k_{\text{idle}}$; active stride $k_{\text{act}}\le k_{\text{idle}}$
\State $t\gets1$
\While{frames remain}
  \State $y\gets[\,s(I_t)>\theta\,]$ \Comment{detector call}
  \State $k\gets k_{\text{act}}$ \textbf{if} $y$ \textbf{else} $k_{\text{idle}}$
  \State $\textsc{Trig}(I_{t}),\dots,\textsc{Trig}(I_{t+k-1})\gets y$ \Comment{hold}
  \State $t\gets t+k$
\EndWhile
\end{algorithmic}
\end{algorithm}

\begin{algorithm}[t]
\caption{Tracking with sparse detections and coasting}
\label{alg:sparse}
\begin{algorithmic}[1]
\Require detections every $k$-th frame; memory $M$; confirmation hits $h$; coasting $c$
\State $\mathcal{K}\gets\emptyset$ \Comment{active tracks}
\For{$t=1,2,\dots$}
  \State predict all tracks in $\mathcal{K}$ \Comment{Kalman or last box}
  \If{$(t-1)\bmod k=0$}
    \State match $D_t$ to predictions by IoU $\ge0.3$
    \State update matched tracks; reset their misses
    \State add a miss to each unmatched track
    \State start a track per unmatched detection
    \State delete tracks with $>\lceil M/k\rceil$ misses
  \EndIf
  \State \textbf{output} tracks with $\ge h$ hits, $\le c$ misses
\EndFor
\end{algorithmic}
\end{algorithm}

\paragraph{Recording policy.}
Frame $t$ is stored if any trigger fires in $[t-2\,\mathrm{s},\, t+1\,\mathrm{s}]$: a 1\,s pre-roll is held in a ring buffer so that the approach of a person is kept, and a 2\,s post-roll avoids cutting clips when the trigger flickers. All methods share this policy (\cref{alg:record}), so storage differences come only from the trigger.

\paragraph{Triggers.}
\emph{Frame differencing (FD):} fraction of pixels whose grey level changes by more than 25 between consecutive blurred frames, after a $3\times3$ opening. \emph{Optical flow:} fraction of pixels with Farneb\"ack~\cite{farneback2003two} magnitude above 1\,px at half resolution. \emph{Background subtraction:} foreground fraction from MOG2~\cite{zivkovic2004improved}, KNN~\cite{zivkovic2006efficient}, CNT and GSOC (OpenCV contrib~\cite{opencv}); shadows count as background. \emph{Detector:} highest confidence among person, bicycle, car, motorcycle, bus and truck from YOLO11n~\cite{yolo11} at 640\,px. \emph{Strided detector:} the detector runs on every $k$-th frame ($k{=}2,4,8$) and its decision is held for $k{-}1$ frames. \emph{Hybrid:} a motion score (MOG2, CNT or FD) opens a gate; the detector runs only on gated frames, and the trigger fires only when both agree. \emph{Adaptive stride} (\cref{alg:adaptive}): the detector runs every $k_{\text{idle}}$ frames while it detects nothing and every $k_{\text{act}}$ frames while an object is present, holding each decision until the next call. We also test the OR of the strided detector and a motion trigger, and a \emph{persistence} filter that fires only if a motion trigger fired in at least $m$ of the last $n$ frames.

\paragraph{Tracking.}
Detections are linked with SORT~\cite{bewley2016sort} (reference implementation; IoU 0.3, 3 hits to confirm). We compare it with the IOU tracker~\cite{bochinski2017iou} and with SORT whose Kalman prediction is replaced by the last matched box, which isolates the motion model, and we vary how long a lost track is kept (\texttt{max\_age} 1, 5, 15). For RQ6, a SORT-style tracker (\cref{alg:sparse}) predicts on every frame but receives detections only on every $k$-th frame, or with each detection dropped independently with probability $p$ (five seeds); \emph{coasting} keeps reporting a confirmed track on its prediction for up to $c$ frames without a matching detection.

\section{Experimental Setup}
\label{sec:setup}

All experiments use public data and a 2-core Intel Xeon CPU at 2.8\,GHz without a GPU (OpenCV 4.10, PyTorch 2.14, Ultralytics 8.4.190, ONNX Runtime 1.31). Latency is the median of isolated runs with 2 threads.

\begin{figure*}[t]
\centering
\includegraphics[width=\textwidth]{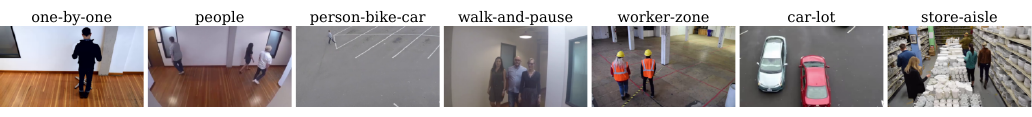}
\caption{The seven benchmark scenes (Intel sample videos~\cite{intelsamples}, CC BY 4.0). \emph{store-aisle} is a busy control clip in which little storage can be saved.}
\label{fig:scenes}
\end{figure*}

\begin{figure}[t]
\centering
\includegraphics[width=\columnwidth]{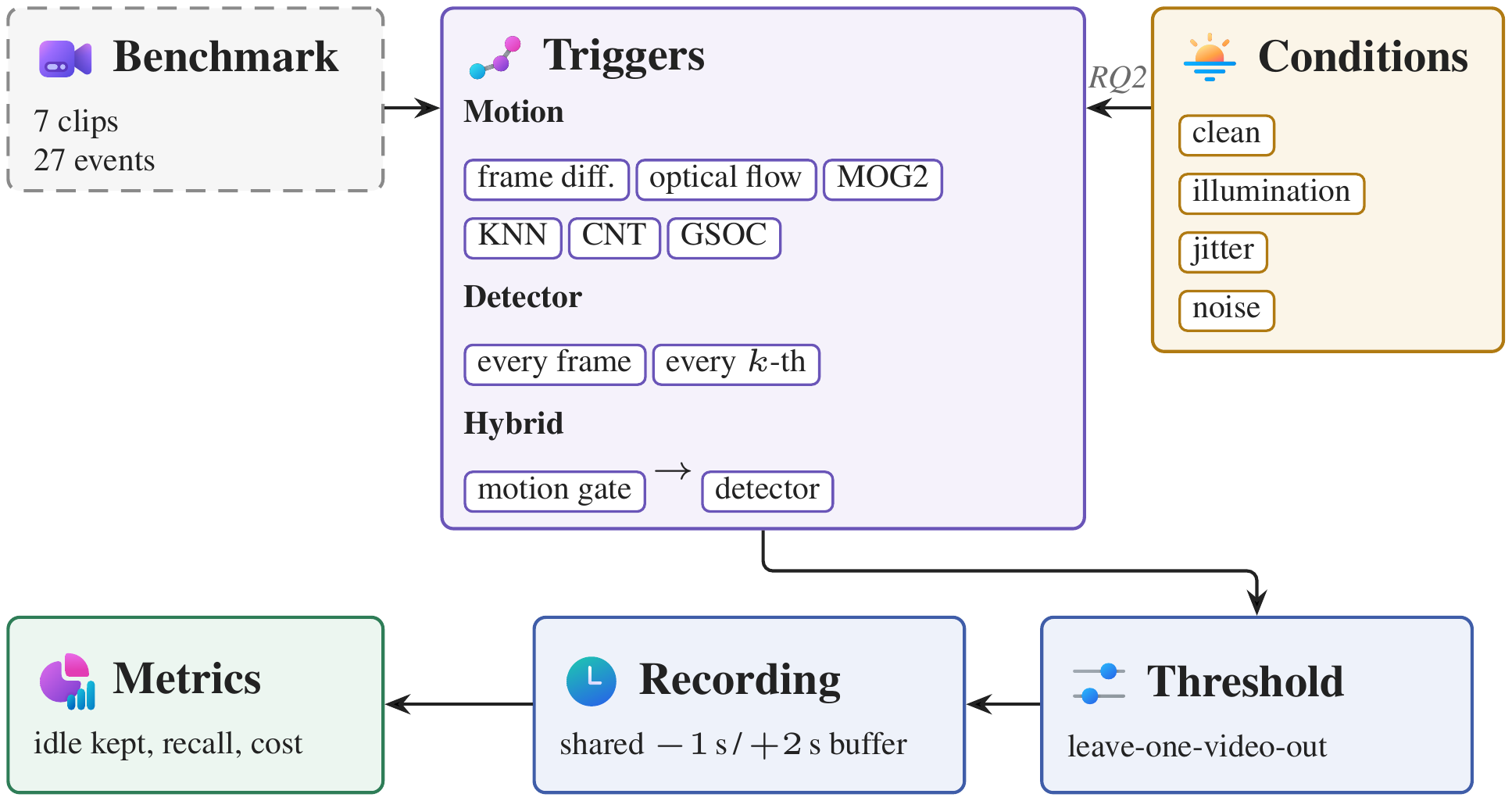}
\caption{Event-recording protocol (RQ1, RQ2). Every trigger is scored on the same clips, thresholded by leave-one-video-out, passed through the same recording buffer, and re-run under three perturbations with thresholds fixed from clean video.}
\label{fig:proto_events}
\end{figure}

\paragraph{Event-recording benchmark.}
We built the benchmark from seven static-camera clips~\cite{intelsamples} (\cref{fig:scenes}), subsampled to about 12\,fps and resized to 768\,px width: 5,800 frames, 27 events, 68.1\% of frames active (per clip 36--90\%). An event is an interval in which at least one person or vehicle is at least partly visible. Intervals were marked at one-second resolution from contact sheets of one frame per second, without reference to any detector output. Frames within 1\,s of an annotated boundary and the first 2\,s of each clip (background warm-up) are excluded from frame-level metrics.

\paragraph{Metrics.}
\emph{Stored}: share of all frames written. \emph{Idle kept}: share of event-free frames stored, the waste a trigger should remove. \emph{Frame recall}: share of event frames stored. \emph{Events captured}: events with at least 90\% of their frames stored. We report 95\% bootstrap intervals over clips.

\paragraph{Protocol.}
Thresholds are set by leave-one-video-out (\cref{fig:proto_events}): for each held-out clip we choose the threshold that stores least on the other six while keeping frame recall $\geq 95\%$ on every one of them, then apply it unchanged (\cref{alg:lovo}). This mimics installing a camera in an unseen scene. As an upper bound we also choose each threshold on the test clip itself (\emph{oracle}).

\paragraph{Perturbations.}
With the same thresholds, every trigger is re-run on three seeded copies of each clip: \emph{illumination}, a global gain that jumps to a value in $[0.6, 1.4]$ every 4--10\,s plus a 10\% drift; \emph{jitter}, a random shift of up to 4\,px per frame; \emph{noise}, Gaussian noise with $\sigma{=}10$ grey levels.

\begin{figure}[t]
\centering
\includegraphics[width=\columnwidth]{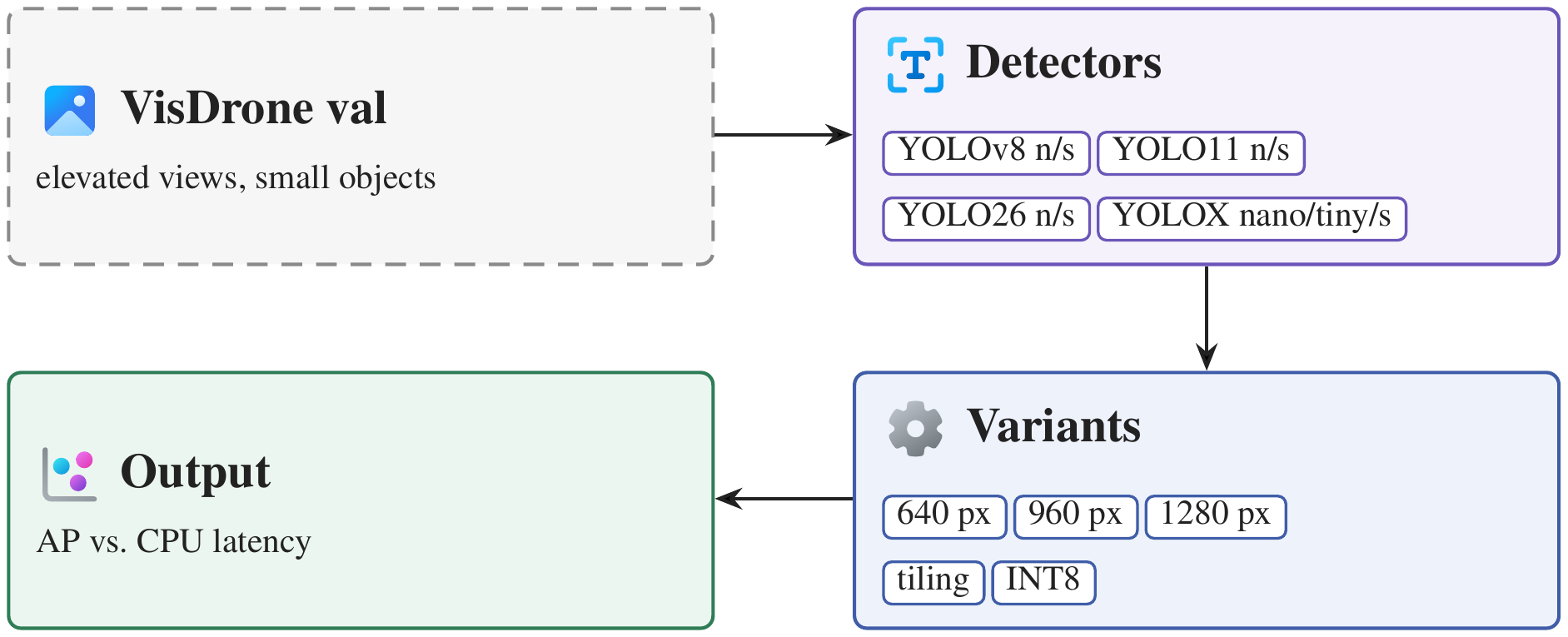}\\[4pt]
\includegraphics[width=\columnwidth]{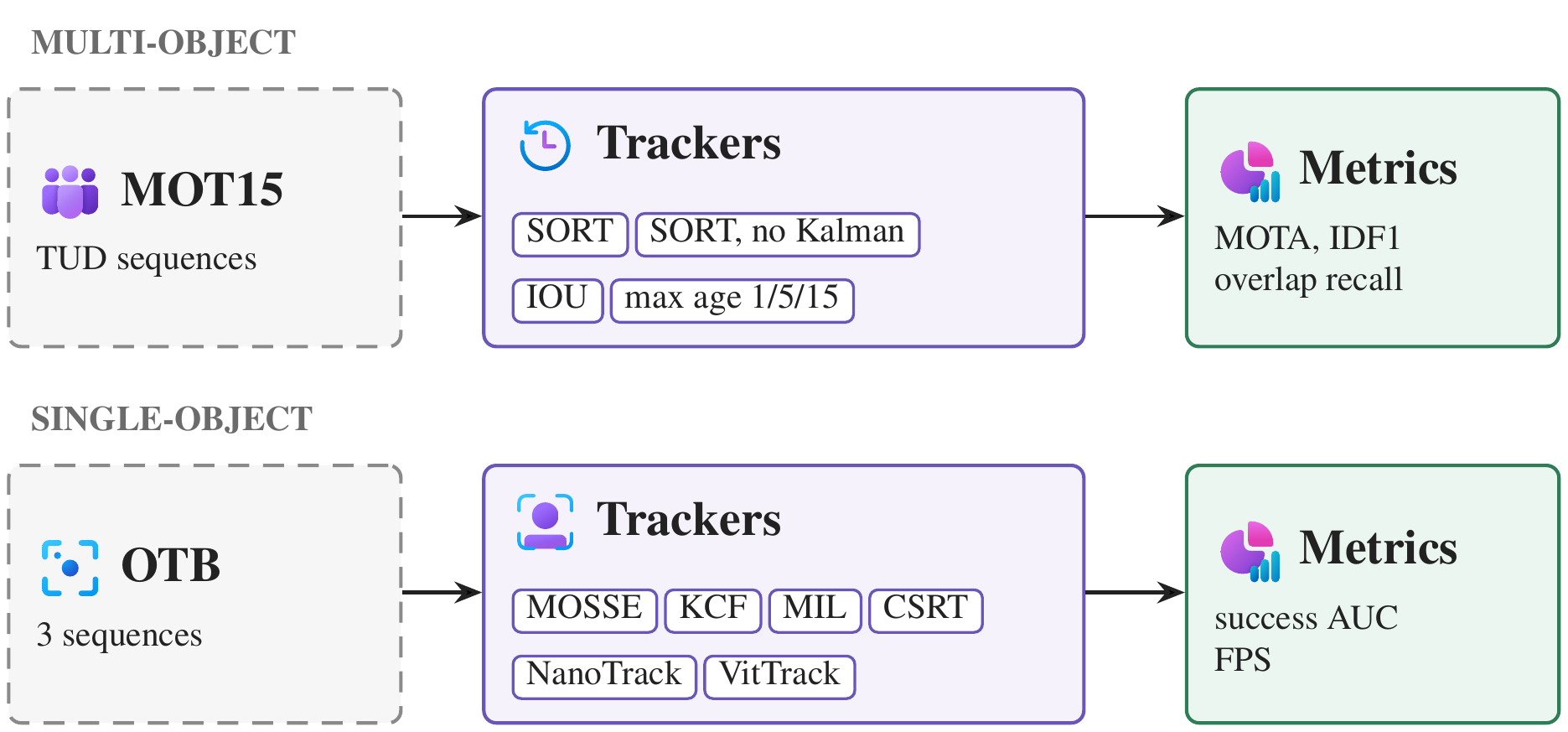}
\caption{Detection (top, RQ3) and tracking (bottom, RQ4) protocols. Detectors are COCO-pretrained and evaluated zero-shot; trackers use the public MOT15 detections or the OTB first-frame box.}
\label{fig:proto_det_trk}
\end{figure}

\paragraph{Detection, quantization and tracking.}
COCO-pretrained detectors are evaluated zero-shot on all 548 VisDrone2019-DET validation images~\cite{zhu2021visdrone} (\cref{fig:proto_det_trk}): pedestrian and people map to \emph{person}; car, van, truck and bus to \emph{vehicle}; ignored regions and ``others'' are crowd regions. We report COCO-style AP with up to 500 detections. INT8 quantization uses ONNX Runtime static per-channel QDQ quantization calibrated on 50 images that are excluded from evaluation. Multi-object tracking uses MOT15 public detections and ground truth for TUD-Campus and TUD-Stadtmitte~\cite{andriluka2008people,lealtaixe2015motchallenge} (250 frames, 1,515 boxes, 18 people) with py-motmetrics at IoU 0.5; a box is \emph{overlapped} when another person's box covers at least 30\% of it. Single-object tracking uses the OTB~\cite{wu2013otb} sequences faceocc2, david and dudek as packaged with OpenCV; the 292 faceocc2 frames the package lists as unsuitable for initialisation, which are the stretches where a book covers the face, form an occlusion subset.

\section{Results}
\label{sec:results}

\subsection{Storage and recall on clean video (RQ1)}

\begin{table}[t]
\centering
\footnotesize
\setlength{\tabcolsep}{3pt}
\begin{tabular}{@{}lccccr@{}}
\toprule
Trigger & Stored & Recall & Events & Det.\ runs & ms/frame \\
\midrule
Frame differencing & 81.5 & 93.6 & 22/27 & 0 & 0.9 \\
Optical flow & 95.2 & 97.5 & 25/27 & 0 & 25.1 \\
MOG2 & 88.2 & 97.6 & 26/27 & 0 & 7.9 \\
KNN & 88.6 & 97.3 & 25/27 & 0 & 7.8 \\
CNT & 86.3 & 97.3 & 24/27 & 0 & 1.4 \\
GSOC & 83.2 & 98.7 & 26/27 & 0 & 32.2 \\
\midrule
Detector, every frame & 79.2 & 99.2 & 26/27 & 100 & 78.7 \\
Detector, every 2nd & 79.2 & 99.2 & 26/27 & 50 & 39.3 \\
Detector, every 4th & 79.5 & 99.0 & 25/27 & 25 & 19.7 \\
\textbf{Detector, every 8th} & 80.2 & \textbf{99.7} & 26/27 & \textbf{12.5} & \textbf{9.8} \\
\midrule
MOG2 gate + detector & 79.1 & 97.6 & 26/27 & 71.0 & 63.8 \\
CNT gate + detector & 77.7 & 98.7 & 25/27 & 78.4 & 63.1 \\
FD gate + detector & 78.4 & 93.6 & 22/27 & 55.8 & 44.8 \\
\bottomrule
\end{tabular}
\caption{Clean video, leave-one-video-out thresholds. Stored, recall and detector runs in \%. CPU cost for gated and strided triggers is the gate cost plus the detector's 78.7\,ms times the share of frames on which it runs.}
\label{tab:events}
\end{table}

\begin{figure}[t]
\centering
\includegraphics[width=\columnwidth]{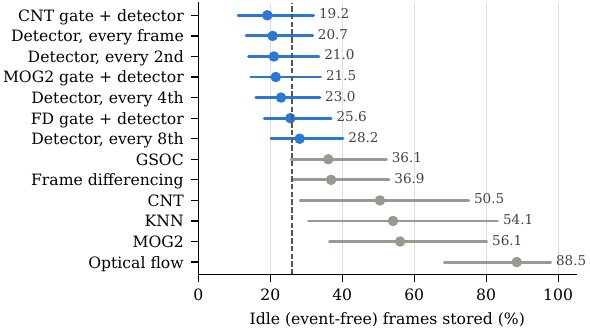}
\caption{Idle frames stored on clean video (blue: uses the detector; grey: motion only). Whiskers: 95\% bootstrap interval over clips. Dashed line at 26.1\%: storing exactly the annotated intervals with the same buffer.}
\label{fig:idle}
\end{figure}

Running the detector on every 8th frame gave the best balance: 99.7\% frame recall, 26 of 27 events captured, and 9.8\,ms of CPU per frame (\cref{tab:events}). Motion triggers stored 36--89\% of idle frames, against 19--28\% for every trigger that used the detector (\cref{fig:idle}). Total stored share varies less (77.7--95.2\%) because 68\% of frames contain an event and the buffer adds about 3\,s around each; storing exactly the annotated intervals with the same buffer would keep 83.1\% of frames and 26.1\% of idle frames. Running the detector less often barely changed storage or recall, because people stay in view for many frames. The detector generation mattered little: with YOLOv8n, YOLO11n and YOLO26n as the trigger, idle storage was 19.0\%, 20.7\% and 23.7\%. The intervals in \cref{fig:idle} are wide because the benchmark has only seven clips.

\subsection{Robustness (RQ2)}

\begin{figure}[t]
\centering
\includegraphics[width=\columnwidth]{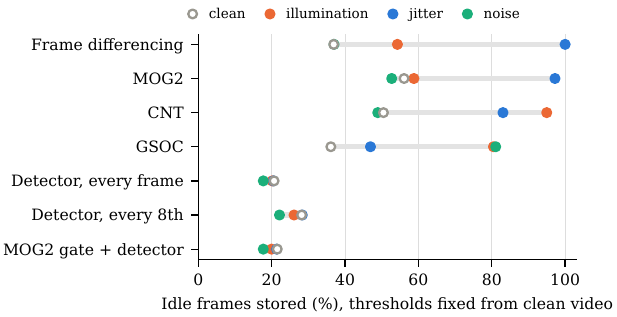}
\caption{Idle frames stored under each perturbation, with thresholds fixed from clean video.}
\label{fig:perturb}
\end{figure}

Camera jitter of up to 4\,px made frame differencing store 100\% of idle frames and MOG2 97.2\%, removing their storage benefit; illumination steps pushed CNT to 95.0\% and GSOC to 80.5\% (\cref{fig:perturb}). Detector-based triggers stayed between 17.2\% and 28.3\% under every condition. Frame recall stayed at or above 93.4\% for all methods and conditions (detector: $\geq 97.7\%$), so the cost of a fragile trigger is storage, not missed events. Gating also lost its compute saving: the MOG2 gate opened on 71.0\% of frames on clean video and 83.2\% under jitter, and the FD gate on 98.9\% under jitter, whereas a strided detector runs on 12.5\% of frames in every condition.

\paragraph{Failure modes.}
Motion triggers lose people who stand still: MOG2 absorbed a person standing at a table and stored only 74\% of that 27\,s event. Detectors fail on viewpoints COCO rarely shows: YOLO11n missed a car seen from directly above in the darker part of \emph{car-lot}, so that 4\,s event was not stored at all.

\subsection{Detector accuracy and CPU cost (RQ3)}

\begin{figure}[t]
\centering
\includegraphics[width=\columnwidth]{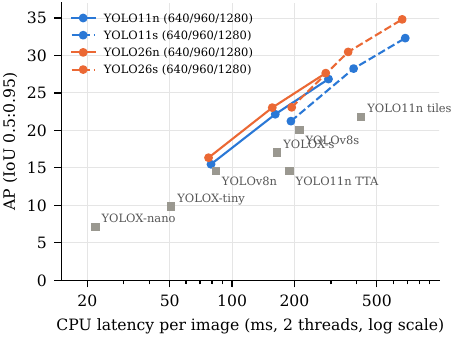}
\caption{Accuracy versus CPU latency on VisDrone2019-DET val. Blue: YOLO11, orange: YOLO26; solid lines are the nano and dashed lines the small models at 640/960/1280\,px. Grey: other models and inference variants.}
\label{fig:speedacc}
\end{figure}

\begin{table}[t]
\centering
\footnotesize
\setlength{\tabcolsep}{2.6pt}
\begin{tabular}{@{}llccccc@{}}
\toprule
Detector & Input & AP & AP$_{50}$ & AP$_S$ & AP$_{\text{pers}}$ & AP$_{\text{veh}}$ \\
\midrule
YOLOX-s & 640 & 17.0 & 32.2 & 7.1 & 6.6 & 27.4 \\
YOLOv8n & 640 & 14.6 & 27.7 & 5.2 & 5.9 & 23.3 \\
YOLO11n & 640 & 15.5 & 29.0 & 6.0 & 6.1 & 24.9 \\
YOLO26n & 640 & 16.3 & 30.8 & 6.4 & 6.5 & 26.2 \\
YOLOv8s & 640 & 20.1 & 36.7 & 9.4 & 9.0 & 31.2 \\
YOLO11s & 640 & 21.2 & 38.3 & 10.4 & 9.7 & 32.8 \\
YOLO26s & 640 & 23.1 & 41.6 & 12.1 & 11.0 & 35.2 \\
YOLO11n & 640, tiles & 21.8 & 40.5 & 13.0 & 11.9 & 31.7 \\
YOLO11n & 960 & 22.2 & 40.1 & 12.0 & 11.2 & 33.1 \\
YOLO11n & 1280 & 26.8 & 47.6 & 17.0 & 15.3 & 38.4 \\
YOLO11s & 960 & 28.2 & 49.4 & 17.5 & 15.8 & 40.7 \\
YOLO11s & 1280 & 32.3 & 55.8 & 22.1 & 20.1 & 44.5 \\
\midrule
YOLO26n & 960 & 23.0 & 41.5 & 12.6 & 11.5 & 34.6 \\
YOLO26n & 1280 & 27.6 & 48.8 & 17.2 & 15.5 & 39.8 \\
YOLO26s & 960 & 30.5 & 53.1 & 19.5 & 17.8 & 43.2 \\
YOLO26s & 1280 & \textbf{34.8} & \textbf{59.8} & \textbf{24.3} & \textbf{22.1} & \textbf{47.5} \\
YOLO11n & 640, TTA & 14.6 & 26.9 & 5.7 & 5.5 & 23.7 \\
\bottomrule
\end{tabular}
\caption{Zero-shot COCO-style AP (\%) on VisDrone2019-DET val (person + vehicle). YOLOX-nano/tiny at 416\,px reach 7.1/9.8 AP. ``tiles'': full image plus $2\times2$ overlapping tiles merged by NMS; ``TTA'': Ultralytics test-time augmentation (flips and rescaling).}
\label{tab:det}
\end{table}

On elevated views with small objects, a larger input helped more than a larger model (\cref{fig:speedacc}, \cref{tab:det}): YOLO11n at 960\,px reached 22.2 AP in 162\,ms, while YOLO11s at 640\,px reached 21.2 AP in 192\,ms. At 1280\,px YOLO11n reached 26.8 AP, 73\% more than at 640\,px. Small objects explain the gap: AP$_S$ rose from 6.0 to 17.0, while AP on large objects barely moved (76.1 to 80.3). Tiled inference reached 21.8 AP but took 420\,ms, worse than resizing to 960\,px. People were much harder than vehicles for every model (best 20.1 vs.\ 44.5 AP). From YOLOv8 to YOLO26, AP rose by 1.7 points for the nano and 3.0 for the small models. The resolution effect held for the newest generation: YOLO26n rose from 16.3 AP at 640\,px to 27.6 at 1280\,px (283\,ms), and YOLO26s at 1280\,px reached 34.8 AP, the best result, 2.5 points above YOLO11s at a similar cost (663 vs.\ 686\,ms). YOLO26n at 960\,px (23.0 AP, 156\,ms) beat YOLO11s at 640\,px (21.2 AP, 192\,ms). Test-time augmentation lowered YOLO11n's AP from 15.5 to 14.6 while more than doubling latency (189\,ms), so it does not help on these views.

\paragraph{INT8 quantization.}
Static INT8 quantization of YOLO11n cut latency from 73.8 to 61.0\,ms ($-17\%$) and size from 10.7 to 4.6\,MB, for AP 15.6$\to$15.3 (AP$_{50}$ 29.4$\to$28.7) on the 498 held-out images. Quantizing the detection head as well produced no usable detections, because box coordinates and class scores share one output tensor with very different ranges; keeping the head in FP32 fixed this. The speed-up is modest even though the CPU supports AVX-512 VNNI; part of the graph still runs in floating point.

\subsection{Tracking under overlap (RQ4)}

\begin{table}[t]
\centering
\footnotesize
\setlength{\tabcolsep}{2.4pt}
\begin{tabular}{@{}lccccc@{}}
\toprule
Tracker & MOTA & IDF1 & IDSW & Rec.\ ovl. & Rec.\ clear \\
\midrule
IOU tracker & 69.5 & 65.8 & 23 & 44.3 & 94.4 \\
No motion model & 67.7 & 71.8 & 18 & 39.6 & 93.1 \\
SORT, \texttt{max\_age}=1 & 69.6 & 70.5 & 16 & 39.9 & 93.9 \\
SORT, \texttt{max\_age}=5 & 69.3 & 71.8 & 14 & 39.0 & 93.8 \\
SORT, \texttt{max\_age}=15 & 69.5 & \textbf{76.7} & \textbf{14} & 39.4 & 93.9 \\
\bottomrule
\end{tabular}
\caption{TUD-Campus + TUD-Stadtmitte, MOT15 public detections. Recall (\%) for the 584 overlapped and 931 clear ground-truth boxes. False positives: 62, 54, 37, 37, 37.}
\label{tab:mot}
\end{table}

Longer track memory helped identity most: raising \texttt{max\_age} from 1 to 15 lifted IDF1 from 70.5 to 76.7 (\cref{tab:mot}). The Kalman model mainly removed false positives (54$\to$37) and fragmentations (37$\to$30) relative to the same tracker without motion prediction. MOTA stayed within 67.7--69.6. Overlap is the bottleneck: every tracker recovered only 39--44\% of overlapped boxes, against 93--94\% of clear ones, because the detector rarely fires on a partly hidden person and no association rule can recover a box that was never detected.

\begin{table}[t]
\centering
\footnotesize
\setlength{\tabcolsep}{2.4pt}
\begin{tabular}{@{}lcccccr@{}}
\toprule
Tracker & faceocc2 & david & dudek & Mean & Occl. & FPS \\
\midrule
MOSSE~\cite{bolme2010mosse} & 60.9 & 53.1 & 14.3 & 42.8 & 66.6 & 1618 \\
KCF~\cite{henriques2015kcf} & 70.4 & 8.6 & 72.3 & 50.4 & 66.8 & 193 \\
MIL~\cite{babenko2011mil} & \textbf{71.4} & 49.8 & 68.7 & 63.3 & \textbf{73.9} & 21 \\
CSRT~\cite{lukezic2017csrdcf} & 68.3 & \textbf{74.2} & 74.1 & 72.2 & 70.3 & 22 \\
NanoTrack (Siamese) & 68.0 & 72.1 & 82.8 & \textbf{74.3} & 63.8 & 67 \\
VitTrack & 54.3 & 66.2 & \textbf{83.3} & 67.9 & 46.5 & 47 \\
\bottomrule
\end{tabular}
\caption{Single-object tracking, OTB one-pass success AUC (\%); ``Occl.'': the 292 occluded faceocc2 frames; FPS on one CPU thread (OpenCV implementations~\cite{opencv}).}
\label{tab:sot}
\end{table}

\paragraph{Single-object trackers under occlusion.}
The Siamese NanoTrack had the best mean AUC (74.3) at 67\,fps, but on occluded faceocc2 frames its AUC fell to 63.8, below MIL (73.9) and CSRT (70.3); the transformer-based VitTrack fell to 46.5 (\cref{tab:sot}). Three sequences are too few for a ranking (KCF's failure on david swings its mean), but on this test occlusion hurt the learned-similarity trackers more than the online-learning ones.

\subsection{Making the trigger cheaper (RQ5)}
\label{sec:rq5}

\begin{figure*}[t]
\centering
\includegraphics[width=\textwidth]{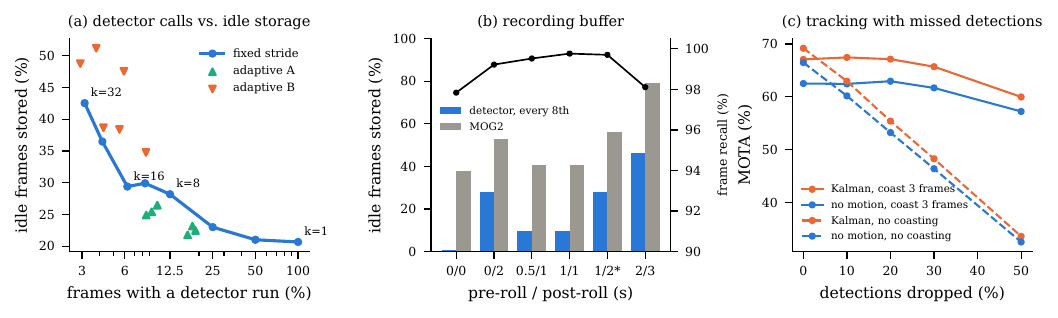}
\caption{(a) Detector calls versus idle storage on clean video for fixed strides ($k{=}1$ to 32) and the two adaptive policies (A: sparse while empty, dense while occupied; B: the reverse). (b) Effect of the pre-/post-roll length (s; * marks the default) on idle storage and, for the strided detector, frame recall. (c) MOTA on TUD when each detection is dropped with the given probability (mean of five seeds), with and without three frames of coasting.}
\label{fig:extended}
\end{figure*}

\begin{table}[t]
\centering
\footnotesize
\setlength{\tabcolsep}{2.6pt}
\begin{tabular}{@{}lccccc@{}}
\toprule
 & \multicolumn{3}{c}{Clean} & Det. & Idle kept \\
\cmidrule(lr){2-4}
Trigger & Idle kept & Recall & Events & runs & jitter / illum. \\
\midrule
Fixed stride 8 & 28.2 & 99.7 & 26/27 & 12.5 & 28.3 / 26.1 \\
Fixed stride 16 & 29.4 & 99.4 & 26/27 & 6.2 & 29.6 / 29.4 \\
Fixed stride 32 & 42.5 & 98.3 & 25/27 & 3.1 & 42.5 / 42.5 \\
\textbf{Adaptive A, 16/8} & \textbf{26.4} & 99.5 & 26/27 & \textbf{10.2} & 26.6 / 24.4 \\
Adaptive A, 24/8 & 25.4 & 98.9 & 26/27 & 9.3 & 26.4 / 25.4 \\
Adaptive A, 16/4 & 23.2 & 99.6 & 26/27 & 18.0 & 23.9 / 23.2 \\
Adaptive B, 8/16 & 34.7 & 99.7 & 26/27 & 8.5 & 35.6 / 32.5 \\
\midrule
Stride 8 OR MOG2 & 21.5 & 96.6 & 24/27 & 12.5 & 22.1 / 41.0 \\
Stride 8 OR CNT & 29.7 & 96.6 & 24/27 & 12.5 & 32.2 / 79.9 \\
\midrule
MOG2 & 56.1 & 97.6 & 26/27 & 0 & 97.2 / 58.8 \\
MOG2, persistence 3/5 & 54.8 & 97.1 & 26/27 & 0 & 86.9 / 59.4 \\
FD & 36.9 & 93.6 & 22/27 & 0 & 100 / 54.3 \\
FD, persistence 3/5 & 37.0 & 90.3 & 21/27 & 0 & 100 / 37.2 \\
\bottomrule
\end{tabular}
\caption{Cheaper and more robust triggers (\%, leave-one-video-out thresholds, 1\,s pre-roll and 2\,s post-roll). Adaptive A uses a sparse stride while the scene is empty and a dense one while an object is present ($k_{\text{idle}}/k_{\text{act}}$, \cref{alg:adaptive}); B does the reverse. OR rules record if either cue fires; persistence fires only if the raw trigger fired in at least 3 of the last 5 frames.}
\label{tab:trig2}
\end{table}

\paragraph{Stride and adaptive stride.}
Fixed strides beyond 8 frames saved detector calls but stored more idle video (\cref{tab:trig2}, \cref{fig:extended}a): at $k{=}16$ the detector ran on 6.2\% of frames and kept 29.4\% of idle frames, at $k{=}32$ on 3.1\% with 42.5\%, because a decision held for 32 frames (2.7\,s) extends each clip well past the end of its event. The adaptive trigger (\cref{alg:adaptive}) avoids this by checking an empty scene sparsely and an occupied one densely, so that clips end promptly. With strides 16/8 it ran the detector on 10.2\% of frames, 18\% fewer than the fixed stride of 8, and kept 26.4\% instead of 28.2\% of idle frames, at 99.5\% recall and with the same 26 of 27 events; with 16/4, idle storage fell to 23.2\% for 18.0\% detector calls. The reverse policy (B), dense when empty and sparse when occupied, was worse than a fixed stride at every setting (34.7\% idle at 8.5\% calls), because long holds after a positive detection lengthen every clip. Both policies kept the detector's robustness: idle storage under jitter and illumination change stayed within 2.2 points of clean video.

\paragraph{Fusion and persistence.}
Combining cues did not help. OR-ing the strided detector with MOG2 lowered idle storage to 21.5\%, but recall fell to 96.6\% and two more events were lost, because the threshold search raised the detector threshold and leaned on MOG2, which then stored 41.0\% of idle frames under illumination change. A persistence filter (3 of the last 5 frames) reduced MOG2's idle storage under jitter from 97.2\% to 86.9\% and frame differencing's under illumination change from 54.3\% to 37.2\%, but it left frame differencing at 100\% under jitter, because shake moves pixels on every frame.

\paragraph{Recording buffer.}
The 2\,s post-roll is the main source of idle storage for the detector trigger (\cref{fig:extended}b). Shortening it to 1\,s cut total storage from 80.2\% to 77.2\% at 99.8\% frame recall with all 26 events; idle kept fell from 28.2\% to 9.5\%, but this overstates the gain, because frames within 1\,s of an annotated boundary are excluded from that metric. Removing the buffer entirely kept only 97.8\% of event frames and 24 events, and a 2\,s/3\,s buffer raised idle storage to 46.1\% for the detector and 79.0\% for MOG2.

\paragraph{Threshold transfer.}
Choosing each threshold on the test clip itself lowered idle storage only from 28.2\% to 22.5\% for the strided detector but from 56.1\% to 39.8\% for MOG2, so motion thresholds transfer to unseen scenes far worse than detector confidence. Even with an oracle threshold, frame differencing could not reach 95\% recall on every clip (92.5\% pooled).

\begin{table}[t]
\centering
\footnotesize
\setlength{\tabcolsep}{2.8pt}
\begin{tabular}{@{}lccc|ccc@{}}
\toprule
 & \multicolumn{3}{c|}{Idle kept (\%)} & \multicolumn{3}{c}{Frame recall (\%)}\\
Trigger & 90 & 95 & 99 & 90 & 95 & 99\\
\midrule
Detector, every 8th & 28.2 & 28.2 & 28.2 & 98.9 & 99.7 & 99.7\\
Adaptive 16/8 & 24.2 & 26.4 & 27.0 & 97.3 & 99.5 & 100.0\\
MOG2 & 55.2 & 56.1 & 58.6 & 95.1 & 97.6 & 97.7\\
CNT & 39.7 & 50.5 & 65.5 & 96.3 & 97.3 & 99.5\\
FD & 36.9 & 36.9 & 36.9 & 93.6 & 93.6 & 93.6\\
\bottomrule
\end{tabular}
\caption{Sensitivity to the recall target $\rho$ of \cref{alg:lovo} (90, 95 or 99\% on every training clip), clean video.}
\label{tab:target}
\end{table}

\paragraph{Recall target.}
The detector was insensitive to the recall target $\rho$ of \cref{alg:lovo}: at $\rho$ of 90\%, 95\% and 99\% the strided detector kept 28.2\% of idle frames in all three cases (\cref{tab:target}), because its confidence separates occupied from empty frames with a wide margin. Motion triggers were not: CNT kept 39.7\%, 50.5\% and 65.5\% of idle frames, and MOG2 55.2--58.6\%, so a stricter recall requirement costs storage mainly for motion cues. Frame differencing never reached 95\% recall on every training clip, so its threshold did not change.

\begin{table}[t]
\centering
\footnotesize
\setlength{\tabcolsep}{3pt}
\begin{tabular}{@{}lccccc@{}}
\toprule
Trigger & Median & 90th pct. & $\le$1\,s & Start & Never \\
 & delay (s) & delay (s) & (\%) & stored (\%) & fired \\
\midrule
Detector, every frame & 0.40 & 0.88 & 89 & 96 & 1 \\
Detector, every 8th & 0.70 & 1.17 & 81 & 93 & 1 \\
Detector, every 16th & 1.03 & 1.84 & 48 & 89 & 1 \\
Adaptive 16/8 & 1.00 & 1.51 & 48 & 93 & 1 \\
\midrule
Optical flow & 0.00 & 0.15 & 100 & 100 & 0 \\
FD & 0.05 & 0.42 & 100 & 100 & 0 \\
MOG2 & 0.02 & 0.50 & 93 & 100 & 0 \\
CNT & 0.02 & 1.21 & 81 & 96 & 1 \\
GSOC & 0.03 & 0.71 & 93 & 96 & 1 \\
\bottomrule
\end{tabular}
\caption{Onset delay on clean video (27 events, leave-one-video-out thresholds): time from the start of an event (annotation tolerance included) to the first trigger. ``$\le$1\,s: share of events whose first trigger falls within the 1\,s pre-roll; ``Start stored: share of events whose frame 1\,s after the annotated start is recorded.}
\label{tab:onset}
\end{table}

\paragraph{Reaction time.}
Storage is not the only cost of a sparse trigger: it also reacts later (\cref{tab:onset}). Motion triggers fired a median of 0.00--0.05\,s after an event began, the detector on every frame after 0.40\,s, every 8th frame after 0.70\,s, and the adaptive trigger after 1.00\,s, because an empty scene is checked only every 16th frame (1.3\,s). The 1\,s pre-roll hides most of this delay: the frame 1\,s into each event was stored for 93\% of events with the strided and adaptive triggers and 96\% with the per-frame detector. Delays longer than the pre-roll do lose the first moments of an event: with the adaptive trigger, 52\% of events were first detected more than 1\,s after they began, against 19\% for the fixed stride of 8. The pre-roll should therefore be at least as long as the idle stride, which for 16 frames at 12\,fps means about 1.5\,s.

\begin{table}[t]
\centering
\footnotesize
\setlength{\tabcolsep}{2.4pt}
\begin{tabular}{@{}lccccccc@{}}
\toprule
Trigger & 1-by-1 & people & p-b-c & walk & worker & car-lot & store\\
\midrule
Det., every 8th & 49 & 26 & 19 & 20 & 79 & 33 & 0\\
Adaptive 16/8 & 49 & 26 & 19 & 20 & 79 & 19 & 0\\
Optical flow & 100 & 100 & 87 & 100 & 93 & 53 & 17\\
FD & 67 & 35 & 26 & 23 & 93 & 37 & 17\\
MOG2 & 72 & 24 & 85 & 31 & 100 & 81 & 50\\
CNT & 94 & 11 & 35 & 63 & 100 & 47 & 45\\
GSOC & 62 & 25 & 31 & 23 & 100 & 47 & 14\\
\bottomrule
\end{tabular}
\caption{Idle frames stored (\%) per scene on clean video, leave-one-video-out thresholds. \emph{store-aisle} has almost no idle time.}
\label{tab:scene}
\end{table}

\paragraph{Per-scene behaviour.}
The ranking of triggers held in most scenes (\cref{tab:scene}), but the spread was large. In \emph{worker-zone}, where machinery and people move continuously, every trigger kept 79--100\% of the few idle frames. MOG2 kept 71--100\% of idle frames in four scenes, and optical flow kept every idle frame in three. Outside \emph{worker-zone} and \emph{store-aisle}, the strided detector kept 19--49\%. With seven clips, per-scene numbers are indicative only.

\subsection{Tracking with sparse or missed detections (RQ6)}
\label{sec:rq6}

\begin{table}[t]
\centering
\footnotesize
\setlength{\tabcolsep}{2.6pt}
\begin{tabular}{@{}clccccc@{}}
\toprule
$k$ & Prediction & MOTA & IDF1 & IDSW & Rec.\ ovl. & Rec.\ clear \\
\midrule
\multirow{2}{*}{1} & Kalman & \textbf{69.2} & \textbf{76.1} & 19 & 42.8 & 93.4 \\
 & last box & 66.5 & 64.8 & 24 & 41.6 & 92.3 \\
\multirow{2}{*}{2} & Kalman & 67.9 & 74.8 & 14 & 41.6 & 89.8 \\
 & last box & 64.3 & 63.1 & 21 & 41.3 & 88.1 \\
\multirow{2}{*}{4} & Kalman & 60.9 & 73.6 & \textbf{6} & 35.4 & 81.7 \\
 & last box & 54.5 & 59.2 & 15 & 34.6 & 76.5 \\
\multirow{2}{*}{8} & Kalman & 38.2 & 55.5 & 5 & 22.3 & 54.4 \\
 & last box & 33.3 & 46.0 & 11 & 21.9 & 51.2 \\
\bottomrule
\end{tabular}
\caption{Tracking when detections arrive only on every $k$-th frame (\cref{alg:sparse}; TUD-Campus + TUD-Stadtmitte, MOT15 public detections, 15-frame track memory). Recall in \% for overlapped and clear ground-truth boxes. With $k{=}1$ this re-implementation differs from \cref{tab:mot} only in the track-confirmation rule.}
\label{tab:mot2}
\end{table}

Strided detection is cheap for recording, but a tracker fed only every $k$-th frame must bridge the gaps (\cref{tab:mot2}). With $k{=}2$ the Kalman tracker lost 1.3 MOTA and 1.3 IDF1 points; with $k{=}4$ MOTA fell to 60.9 while IDF1 stayed at 73.6 and identity switches dropped from 19 to 6; with $k{=}8$ recall of clear boxes halved to 54.4\%, because boxes predicted over eight frames drift away from walking people. The Kalman model mattered more as detections became sparser: its MOTA lead over last-box prediction grew from 2.7 points at $k{=}1$ to 6.4 at $k{=}4$, and its IDF1 lead was 9--14 points at every stride.

Random detection dropout shows the role of coasting (\cref{fig:extended}c). Without coasting, MOTA fell from 69.2 to 48.3 when 30\% of detections were dropped, because every missed detection hides its track for that frame. With three frames of coasting, the Kalman tracker held 65.7 MOTA, against 61.7 without a motion model, and on the unmodified detections its recall of overlapped people rose from 42.8\% to 49.7\%, at the cost of more false positives (53 to 130). Coasting on a motion model is therefore the cheapest way to recover people who are briefly hidden, which is the failure that limited every tracker in \cref{tab:mot}.

\begin{table}[t]
\centering
\footnotesize
\setlength{\tabcolsep}{3pt}
\begin{tabular}{@{}llcccc@{}}
\toprule
Parameter & Value & MOTA & IDF1 & FP & Rec.\ ovl.\\
\midrule
\multirow{5}{*}{Coasting $c$ (frames)} & 0 & 69.2 & 76.1 & 53 & 42.8\\
 & 1 & 68.8 & 75.6 & 78 & 45.9\\
 & 3 & 67.1 & 74.5 & 130 & 49.7\\
 & 5 & 65.0 & 73.5 & 179 & 52.9\\
 & 10 & 58.3 & 70.7 & 302 & 56.7\\
\midrule
\multirow{4}{*}{Confirmation $h$ (hits)} & 1 & 63.4 & 72.3 & 239 & 56.5\\
 & 2 & 65.5 & 73.8 & 177 & 52.6\\
 & 3 & 67.1 & 74.5 & 130 & 49.7\\
 & 5 & 66.7 & 74.7 & 87 & 45.4\\
\midrule
\multirow{4}{*}{IoU gate} & 0.1 & 67.0 & 75.5 & 136 & 50.5\\
 & 0.2 & 66.5 & 74.4 & 139 & 50.0\\
 & 0.3 & 67.1 & 74.5 & 130 & 49.7\\
 & 0.5 & 66.4 & 71.5 & 117 & 47.1\\
\bottomrule
\end{tabular}
\caption{Sensitivity of the Kalman tracker (\cref{alg:sparse}, $k{=}1$) to its parameters on TUD, MOT15 public detections. Unless varied: $c{=}3$, $h{=}3$, IoU gate 0.3 (the coasting sweep uses $c$ as listed).}
\label{tab:mot3}
\end{table}

\paragraph{Tracker parameters.}
Coasting trades false positives for recall of hidden people (\cref{tab:mot3}): from 0 to 10 frames, recall of overlapped boxes rose from 42.8\% to 56.7\%, but false positives grew from 53 to 302 and MOTA fell from 69.2 to 58.3, because a coasting box is also kept after a person has really left. One to three frames is a reasonable compromise. Requiring fewer hits to confirm a track had the same effect (one hit: 56.5\% overlapped recall, 239 false positives), and the IoU gate mattered little between 0.1 and 0.3 (MOTA 66.5--67.1).

\section{Discussion and Limitations}
\label{sec:discussion}

\paragraph{What to deploy.}
For a static camera on a CPU, run a person-and-vehicle detector on every 8th frame and record with a pre- and post-roll buffer: here this kept 99.7\% of event frames, stored 28.2\% of idle frames, and needed the CNN on 12.5\% of frames in every condition. Motion gating saved fewer detector calls on clean video (29\% for MOG2) and less under shake (17\% for MOG2, 1\% for FD). An adaptive stride (every 16th frame while the scene is empty, every 8th while it is occupied) saves a further 18\% of detector calls and stores slightly less, and a 1\,s post-roll appears sufficient on our clips. If the same detections also feed a tracker, run the detector at least every second frame and let a Kalman tracker coast through short gaps. Deployments should still check their own scenes for stationary people and unusual viewpoints.

\paragraph{Limitations.}
The event benchmark is small (seven clips, 27 events), so intervals are wide; findings should be confirmed on long real CCTV recordings with idle periods of hours. Events were marked at one-second resolution by a single annotator, and frames near boundaries were excluded. Perturbations were applied digitally and do not reproduce night scenes, rain or real auto-exposure, and detection dropout in RQ6 is random, whereas real misses cluster during occlusion. The buffer comparison is limited by the one-second annotation resolution. CDnet 2014, MOT17/MOT20 and pretrained SSD and Faster R-CNN weights were not available in our environment, so the tracking study uses two short MOT15 sequences, appearance-based trackers are not evaluated, and HOTA is not reported. All timings come from one 2-core CPU; absolute numbers will differ on edge devices and GPUs.

\section{Conclusion}
\label{sec:conclusion}

For event-triggered recording on a CPU, a detector is a better trigger than motion, and it need not run on every frame: on our benchmark, YOLO11n on every 8th frame kept 99.7\% of event frames, stored 28\% of idle frames rather than 36--89\%, and cost under 10\,ms per frame. Motion triggers lost their storage benefit under jitter or lighting change, and motion gating saved few detector calls. For elevated views, input resolution mattered more than model size; for tracking, detecting partly hidden people mattered more than association. An adaptive stride makes the trigger cheaper still, and a Kalman tracker with coasting tolerates sparse or missing detections. Future work should use long real-world recordings, appearance-based trackers and MOT17/MOT20, and test adaptive strides on hours of idle footage.

{
    \small

}

\end{document}